# An Algorithm to Effect Prompt Termination of Myopic Local Search on Kauffman's NK Landscape

**Sasanka Sekhar Chanda**

Professor, Strategic Management, Indian Institute of Management Indore
C-101 Academic Block, IIM Indore, Prabandh Shikhar, Rau-Pithampur Road, Indore, MP, India 453556.
Email: sasanka2012@gmail.com    Phone (Office): +91 731 2439 591

---

**Abstract**. In Kauffman's *NK* model, myopic local search involves flipping one randomly-chosen bit of an *N*-bit decision string in every time step and accepting the new configuration if that has higher fitness. One issue is that, this algorithm consumes the full extent of computational resources allocated—given by the number of alternative configurations inspected—even though search is expected to terminate the moment there are no neighbors having higher fitness. Otherwise, the algorithm must compute the fitness of all *N* neighbors in every time step, consuming a high amount of resources. In order to get around this problem, I describe an algorithm that allows search to *logically* terminate relatively early, without having to evaluate fitness of all *N* neighbors at every time step. I further suggest that when the efficacy of two algorithms need to be compared head to head, imposing a common limit on the number of alternatives evaluated—metering—provides the necessary level field.

---

# 1    Introduction

Kauffman's *NK*-model [1]—involving search for high fitness decision-configurations—has been invoked in many disciplines: Physics [2], Biology [1,3], Management [4], Organization Studies [5] Public Administration [6] etc. The *NK*-model formulates a decision situation as one of finding a high fitness decision-configuration by searching on a rugged fitness landscape, where a given decision-configuration has **N** dimensions, and the correlation between overall fitness values at neighboring configurations atrophy when **K**—a measure of interdependence between the decision dimensions—is high (**K** > 3) [7].

In the *NK* model, a decision-configuration is an *N*-bit string. Each bit (or *node* or *dimension*) can have (state) values of "0" or "1". The ***fitness contribution*** by one node depends on its state value ("0" or "1") and on the state values in **K** other nodes on which the focal bit is dependent. The values of such ***fitness contribution***— $f_i$ (**i** varies from 1 to *N*)— are drawn from the *Uniform Distribution*, i.e. $0 < f_i < 1$. Higher the value of **K** (or, in some accounts, higher the value of **K/N**) higher is the *complexity* arising from pervasive interdependence. The ***overall fitness*** of a decision-configuration is computed as the arithmetic average of the fitness contributions by the constituent dimensions.[1]

The number of possible decision-configurations in the *NK*-model is $2^N$. Thus, in order to find the decision-configuration having the highest overall fitness, $2^N$ computations of ***overall fitness*** need to be carried out. As **N** increases, the extent of computing resources necessary to find the decision-configuration with the highest fitness increases exponentially. In reality, computing resources are likely to be limited. In order to attain a decision-configuration having reasonably high ***overall fitness*** with limited computational resources, a *simple model of myopic local search* (hereafter referred to as **SMMLS**) is invoked Knudsen and Levinthal [5]. This algorithm is also referred to as *local search* in Levinthal [4] and as *centralized search* in Siggelkow and Levinthal [8] and Siggelkow and Rivkin [9].

To operate the **SMMLS** algorithm, a starting configuration of an **N**-bit decision string is generated by randomly assigning "0" and "1" with equal probability (0.50). One randomly-chosen bit is flipped in every time step. The new configuration is accepted if that has higher fitness. One fitness computation is carried out in a given time step. Effectively, in the **SMMLS**, the extent of computational resources—i.e. the limit on the number of alternative configurations assessed—equals the number of time steps a simulation experiment is run for.

We note that the **SMMLS** algorithm has no way of *being aware* that a decision-configuration has been reached from where no further moves are possible, i.e., all neighbors that are different from the current decision-configuration by one bit have lower fitness.[2] Due to such lack of awareness, the **SMMLS** will continue to attempt to seek higher fitness configurations by flipping a randomly chosen node in all the remaining time steps, fruitlessly. Alternately, the algorithm may be modified to compute the fitness of all **N** neighbors in every time step. This requires a high amount of resources. In such a case, when the extent of available computational resources (permitted number of computations of ***overall fitness***) is low the algorithm will, perforce, have to terminate before it reaches 'equilibrium'—a condition where all **N** immediate neighbors have lower fitness. Moreover, it is also difficult to justify evaluating all **N** neighbors in any given time step, but making a decision to move or stay based only on the fitness of the first evaluated decision-configuration.

---

[1] Interested readers may refer to Ganco and Hoetker [10] for further details of the *NK*-model.
[2] This state of 'equilibrium', where a decision-configuration has fitness higher than all **N** neighbors (that differ from the focal configuration by exactly one bit-value) is referred to as a *competency trap* by Rivkin and Siggelkow [11], who draw from Levinthal and March [12] and from Levitt and March [13].

The situation described above is sub-optimal. In order to get around this problem, I describe an algorithm—an *intelligent model of myopic local search* (hereafter **IMMLS**) that allows search to *logically* terminate relatively early, without having to evaluate fitness of all **N** neighbors at every time step. I further discuss *algorithm effectiveness* by proposing measures to compare algorithms on a common base.

## 2  Algorithm Comparisons: Metering

We noted above that the **SMMLS** evaluates one alternative decision-configuration in a given time step and runs for **T** time steps. Thus the total number of alternative decision-configurations that can be evaluated is capped at **T**. To ensure a level playing field, any rival algorithm should also be limited to evaluation of a maximum of **T** alternatives. This is in view of the fact that the (proposed) rival algorithm may evaluate more than one alternative in any given time step. In such a case, the (rival) algorithm will, effectively, execute for less than **T** time steps. I designate the practice of limiting the number of alternative decision-configurations evaluated as **metering**. Though they don't use the term **metering**, Siggelkow and Rivkin [9: 115] also suggest bringing algorithms on to a common base by fixing the number of alternative decision-configurations evaluated.

When two algorithms endeavor to attain a common goal by employing different logic / mechanisms, **metering** allows comparison from the standpoint of effectiveness. The practice of **metering** recommends providing the same extent of resources *only* for one big-ticket item in *NK*-model computer programs—calculation of *overall fitness* of a decision-configuration. The concept of **Metering** proposed here ignores potential other differences[3] between the computer codes of two algorithms that could be additionally considered for comparison of algorithm efficacy.

## 3  An Intelligent Model of Myopic Local Search

Similar to the **SMMLS** algorithm, in the *intelligent model of myopic local search* (**IMMLS**) an *initial decision-configuration* or the starting configuration is generated by randomly assigning "0" and "1" with equal probability (0.50) to the *N* nodes of the decision string. Likewise, the number of time steps is set to **T**. In addition, in the **IMMLS**, a *metering parameter* is set to the value **T**. At runtime this parameter is accessed to ensure that the **IMMLS** algorithm carries out no more than **T** computations of *overall fitness*.

---

[3] For example, computer code logic for any given function can be developed by multiple approaches, some using a higher number of computations and less memory, others doing the opposite. The trade-offs between computational intensity and memory usage are kept out of scope of this study.

The **IMMLS** algorithm also differs from the **SMMLS** algorithm in terms of what transpires in any given time step. In the **IMMLS**, in any given time step, decision nodes are flipped in sequence, starting with the first node. If flipping a node does not yield a configuration having higher fitness, the next node in the sequence is flipped. If a configuration having higher fitness is obtained, the move is finalized—i.e. the decision-configuration is updated—and execution proceeds to the next time step. Further, if, in a given time step, flipping all the bits starting from the first bit and ending with the $N^{th}$ bit fails to yield a decision-configuration having overall fitness higher than the overall fitness of the current configuration, the algorithm terminates. Lastly, one unit of resource consumption is recognized on each occasion that the overall fitness of a decision-configuration is evaluated. The **IMMLS** algorithm terminates after the extent of resources consumed reaches the value set in the *metering parameter*. Under conditions of low resources allocation, this termination clause precedes the 'equilibrium'-oriented termination clause that the **IMMLS** quits searching if no node can be flipped upon checking all **N** nodes in sequence.

## 4 Results

In this section, first I describe the effort with respect to docking of the *NK*-model code with prior results available in the public domain. Next, I describe the experiments done with the **SMMLS** and **IMMLS** algorithms. I also explain some minor differences in outcomes by delving into model mechanisms. Thereafter I discuss some robustness checks.

### 4.1 Docking of the NK-model code with *Sendero*

I created the *NK*-model on the MATLAB software application platform. The results from runs with the **SMMLS** algorithm were tallied with Kauffman's original results, as reported by the *Sendero* project on its website [14]. In **Table 1** I provide comparative figures. We satisfy ourselves that values obtained do not differ by more than a percentage point from the Kauffman/ *Sendero* results.

**Table 1.** Comparison of **Fitness** values: Kauffman, Sendero and **SMMLS**

|        | *K =>*   | *0*  | *2*  | *4*  | *8*  | *15* |
|--------|----------|------|------|------|------|------|
| *N* = 16 | Kauffman | 0.65 | 0.70 | 0.71 | 0.68 | 0.65 |
|        | Sendero  | 0.67 | 0.71 | 0.70 | 0.68 | 0.64 |
|        | *SMMLS*  | 0.66 | 0.71 | 0.71 | 0.69 | 0.64 |

*Notes.* Fitness values range from zero to one.

### 4.2 Overview of the simulation experiments

I ran the **SMMLS** and **IMMLS** algorithms on **NK** landscapes having **N** = 16 and interdependence varying from 0 to 15 (**N** – 1). Each experiment is run on 10,000 distinct landscapes with a random starting configuration. The results reported are averages over 10,000 experiments.

In Figure 1 I present the differences in *overall fitness* values between **SMMLS** and **IMMLS** algorithms executing for a range of time periods (**T**). As discussed above, the number of alternative decision-configurations evaluated is capped to **T**. The values on the vertical axis are computed as (final overall fitness attained by **SMMLS** - final overall fitness attained by **IMMLS**). Figure 2 shows the main result of this study – conclusively demonstrating **IMMLS**'s ability to exit execution promptly, relative to **SMMLS**. Figures 3, 4, 5 and 6 help understand the (minor) differences between outcomes of **SMMLS** and **IMMLS** observed in Figure 1 by delving into model mechanisms. Figures 7 and 8 highlight some further results, under stylized scenarios of high complexity.

### 4.3 Comparison between outcomes of SMMLS and IMMLS

In **Figure 1** we observe that there is virtually no difference between **SMMLS** and **IMMLS** with respect to the levels of fitness attained for **T**≥100. We are thus assured that switching to **IMMLS** will not compromise the level of outcome attained, while conferring the benefits of earlier exit—lesser computation—relative to **SMMLS**.[4]

**Figure 1.** Difference in overall fitness between **SMMLS** and **IMMLS**

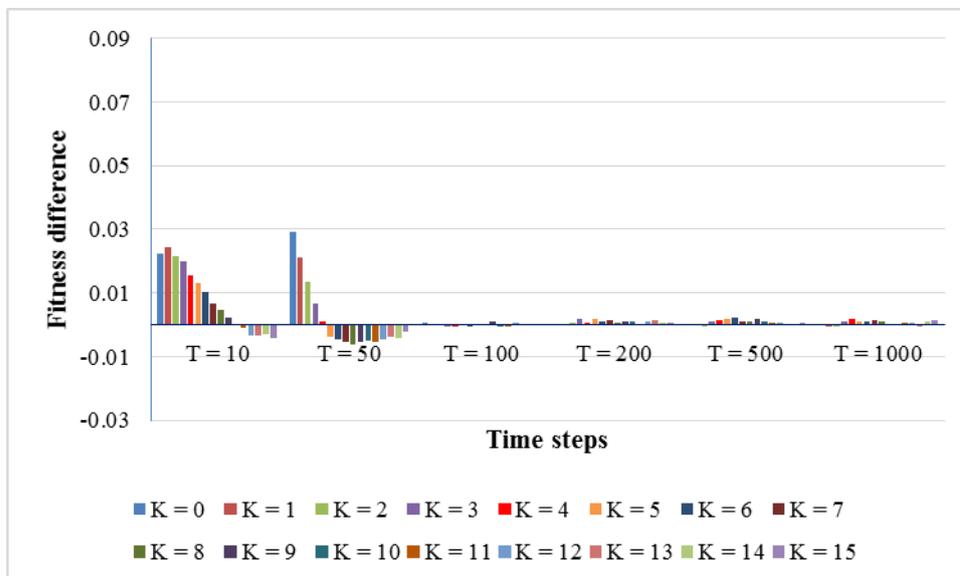

*Parameters.* **N** = 16; Number of iterations = 10,000; *Fitness difference* = (overall fitness by **SMMLS** – overall fitness by **IMMLS**).

In **Figure 2** I present the picture regarding resource consumption, for the situation illustrated in Figure 1. The values in the vertical axis represent the ratio (number of overall fitness computations performed by the **IMMLS** before exiting / **T**), expressed as a percentage. As discussed earlier, **SMMLS** consumes 100% of the resources (given by the value of **T**) in all cases. For **IMMLS**, onwards experiments for *T* = 50, we see a clear pattern comprising two characteristics (a) the extent of resources consumed progressively decrease as **T** increases (b) the extent of resources consumed progressively decrease as interdependence (***K***) or complexity increases. For this reason I recommend **IMMLS** to be the baseline model for search on *NK* landscapes, in place of the **SMMLS**.

---

[4] We note though, the **SMMLS** outperforms the **IMMLS** for **T** = 10 and for **K**<10 and for **K**<5 for **T** = 50. I discuss this later.

**Figure 2.** Resource consumption for **IMMLS** algorithm

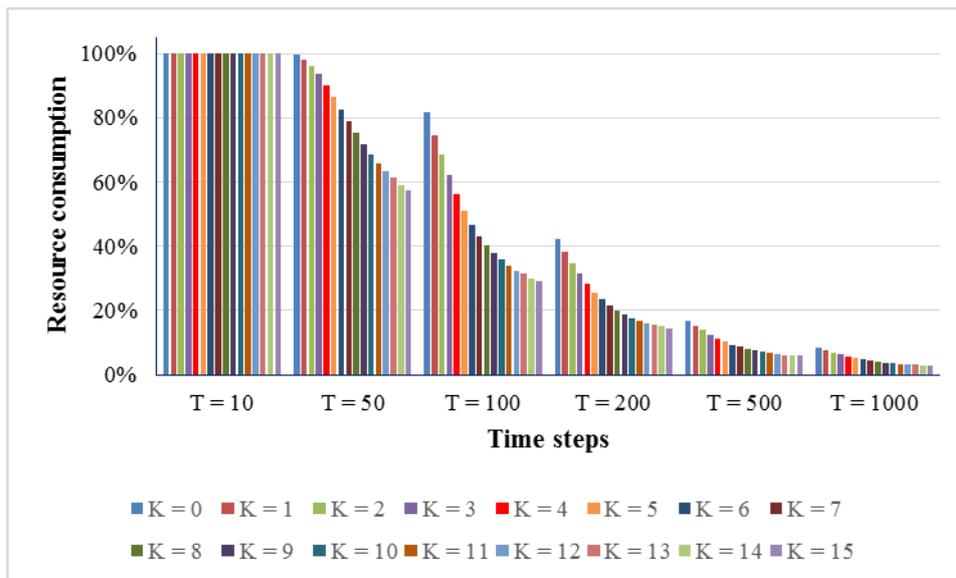

*Parameters.* **N** = 16; Number of iterations = 10,000; ***Resource consumption*** represents the number of alternative configurations evaluated to arrive at the final decision-configuration.

## 4.4 Explanation of the differences between outcomes of SMMLS & IMMLS

We noted in Figure **1** that for low values of the number of periods of simulation **T** (**T** = 10 and **T** = 50) the **SMMLS** obtains higher overall fitness than that obtained by the **IMMLS**, when complexity or *interdependence* **K** (for **K** < 10 and for **K** < 5 respectively) is not high. We also observed that the differences between overall fitness attained vanishes for higher values of **T** (say upwards of **T** = 100). In order to find an explanation as to why **SMMLS** and **IMMLS** obtain somewhat different overall fitness on some occasions, we first calculate the value of the **extent of fitness improvement** attained, as shown below. This metric helps remove idiosyncrasies that might arise from running the experiments on different fitness landscapes, with random starting points.

Fitness Improvement = (Overall fitness at final decision-configuration – Overall fitness at the initial decision-configuration) / Overall fitness at the initial decision-configuration.

Certain prior researchers [9, 11] have espoused an approach of reporting the ratio [final fitness / maximum fitness attainable in a given landscape] as a measure of performance or payoff obtained an algorithm. However, this requires laborious calculation of fitness of all $2^N$ configurations of all 10,000 fitness landscapes, with negligible benefit. Such calculations are also unworkable for large values of **N**. Going forward I recommend using *Fitness Improvement* as a measure of performance or payoff obtained from an algorithm.

**Figure 3.** Comparison of fitness improvement **SMMLS** and **IMMLS**, **T** = 10

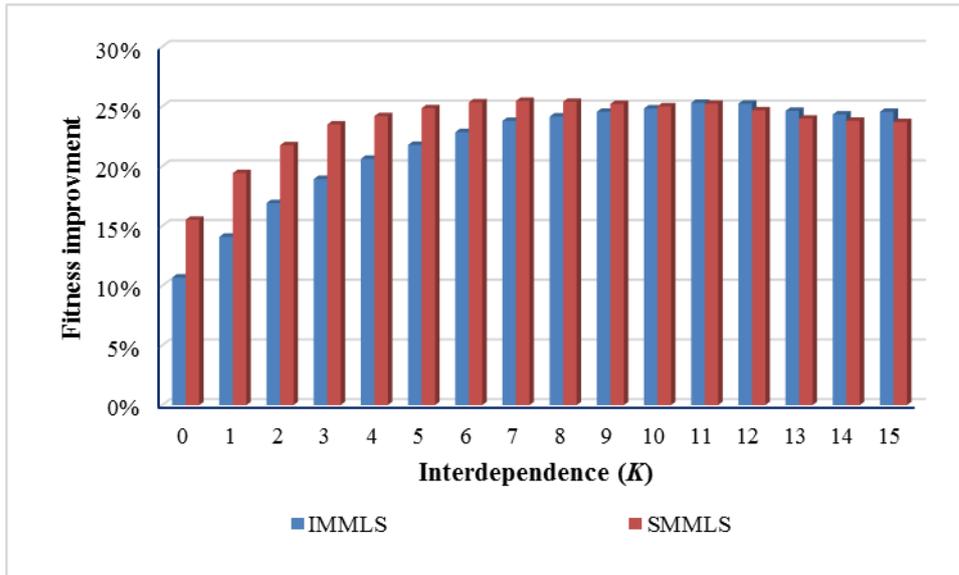

*Parameters.* **N** = 16; **T** = 10; Number of iterations = 10,000; *Fitness improvement* is computed as (Overall fitness at final decision-configuration – Overall fitness at the initial decision-configuration) / Overall fitness at the initial decision-configuration.

**Figure 4.** Comparison of fitness improvement **SMMLS** and **IMMLS**, **T** = 100

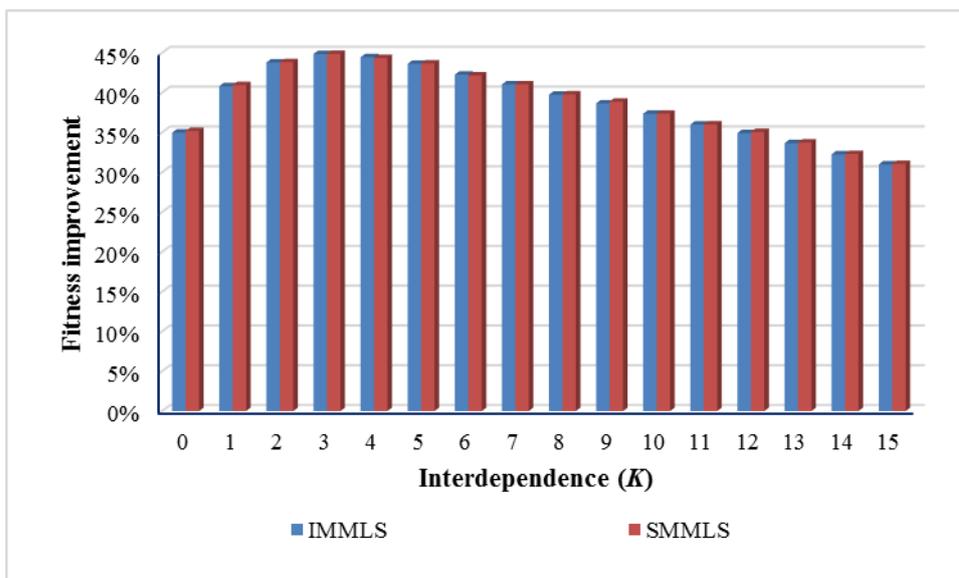

*Parameters.* **N** = 16; **T** = 100; Number of iterations = 10,000; *Fitness improvement* is computed as (Overall fitness at final decision-configuration – Overall fitness at the initial decision-configuration) / Overall fitness at the initial decision-configuration.

In **Figure 3** we compare the **fitness improvement** attained by **SMMLS** and **IMMLS**, for **T** =10. We present the corresponding results for **T** = 100 in **Figure 4**. We observe that **SMMLS** brings about a higher extent of fitness improvement for **K** < 10, when **T** =10 (Figure 3). From Figure 4 we observe that the extent of fitness improvement is comparable, for **T** =100.

**Figure 5.** Comparison of number of successful moves **SMMLS** and **IMMLS**, **T** = 10

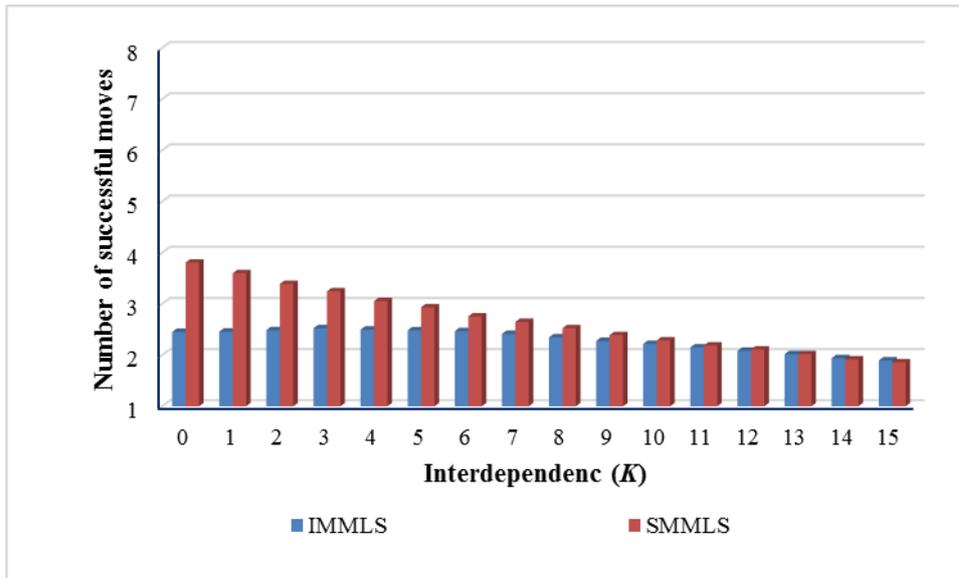

*Parameters.* **N** = 16; **T** = 10; Number of iterations = 10,000; *Number of successful moves* represent the number of times a higher fitness decision-configuration was found and (thereafter) traversed to.

**Figure 6.** Comparison of number of successful moves **SMMLS** and **IMMLS**, **T** = 50

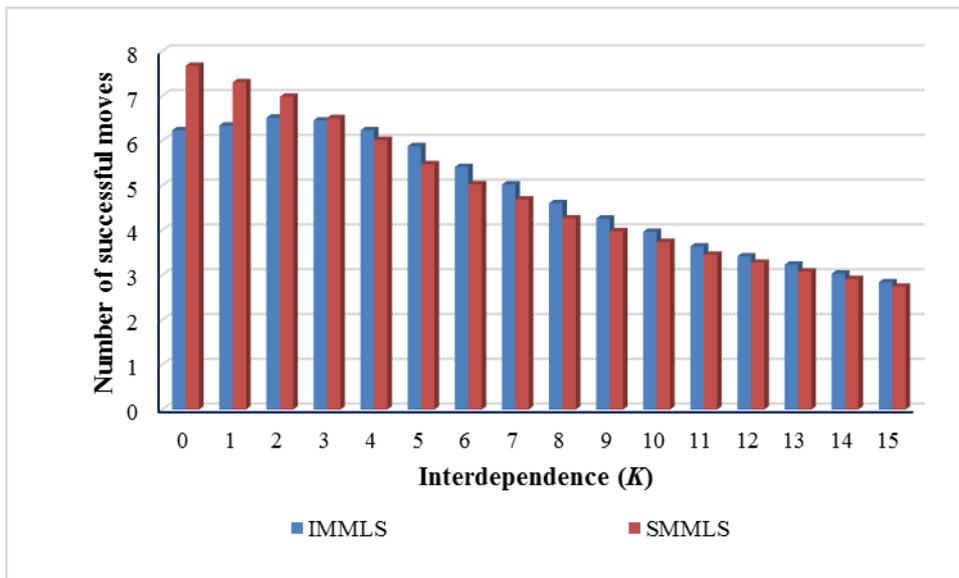

*Parameters.* **N** = 16; **T** = 50; Number of iterations = 10,000; *Number of successful moves* represent the number of times a higher fitness decision-configuration was found and (thereafter) traversed to.

Thus we are left with explaining why **SMMLS** is able to impart higher extent of improvement in fitness—compared to the extent of improvement imparted by **IMMLS**—when **T** = 10 and **K** < 10 and when **T** = 50 and **K** < 5. We hypothesize that this can happen if, for some reason, the **SMMLS** is able to find a higher number of configurations to successively traverse to. To test this intuition, in **Figure 5** we present a comparison of **number of successful moves** by **SMMLS** and **IMMLS**, **T** = 10. We present the corresponding picture for **T** = 50 in **Figure 6**.

The number of successful moves represent the number of times a higher fitness decision-configuration was found and (thereafter) traversed to. This metric pays due regard to the fact that not all attempted flips of a decision-node results in a successful move to a higher fitness configuration. When resources (permitted number of computations of overall fitness of alternative decision-configurations) are limited, an algorithm accomplishing a higher number of successful moves is likely to outperform one accomplishing a lower number of successful moves.

We observe that for **T** = 10 and **K** ≤ 10 (Figure 5) and for **T** = 50 with **K** < 4 (Figure 6), **SMMLS** is indeed able to find a higher number of configurations to successively traverse to. This happens because it randomly picks a bit to flip. In comparison, **IMMLS** is likely to be a bit handicapped, since it keeps flipping nodes in sequence, in any given time step. We further know that landscapes are more correlated when *K* is low [7]. Finally, more correlated a landscape worse will be the outcomes of flipping nodes in sequence (as against flipping nodes selected randomly). Thus **SMMLS** will have a certain extent of advantage for on low and moderate **K** (**K**<10) landscapes for the lowest value of **T** (**T** = 10). When we use **T** = 50, the advantage shrinks to only low **K** (**K** < 4) landscapes.

### 4.5 Robustness checks

Robustness checks inform about the extent of generalizability of the model results. The set of experiments reported above have a very lean set of parameters: **N**, **K** and **T**. Since we already report outcomes under varying **T**, we need to check robustness only for varying **N**.

I carried out similar experimentation for **N** = 20 and **K** running from 0 to 19 and obtained similar pattern of outcomes.[5] Further, upon observing **IMMLS** marginally outperform **SMMLS** for high **K**, I ran a set of experiments varying **N** from 16 to 20 and using **K** = **N** -1 in each case. The rationale for this is that some researchers [15: 1784] consider **K/N** as a more appropriate measure of complexity stemming from pervasive interdependence between the nodes of a decision-configuration (instead of simply using **K**). In **Figure 7** I present the results of experiments with the **IMMLS** algorithm executed with varying **K/N**. Analogous to what was done for Figure 1, I plot the **Fitness Difference** between **SMMLS** and **IMMLS** under varying number of time steps (**T**)—i.e. under varying extent of resources endowed—and for varying complexity (given by **K/N**). We observe that for low values of **T**, **IMMLS** slightly outperforms **SMMLS**, though the difference is never more than half a percentage point.

---
[5] The figures equivalent to Figures 1 and 2 with **N** = 20 (instead of **N** = 16) are provided in the Appendix.

**Figure 7**. Difference in overall fitness between **SMMLS** and **IMMLS** at high complexity

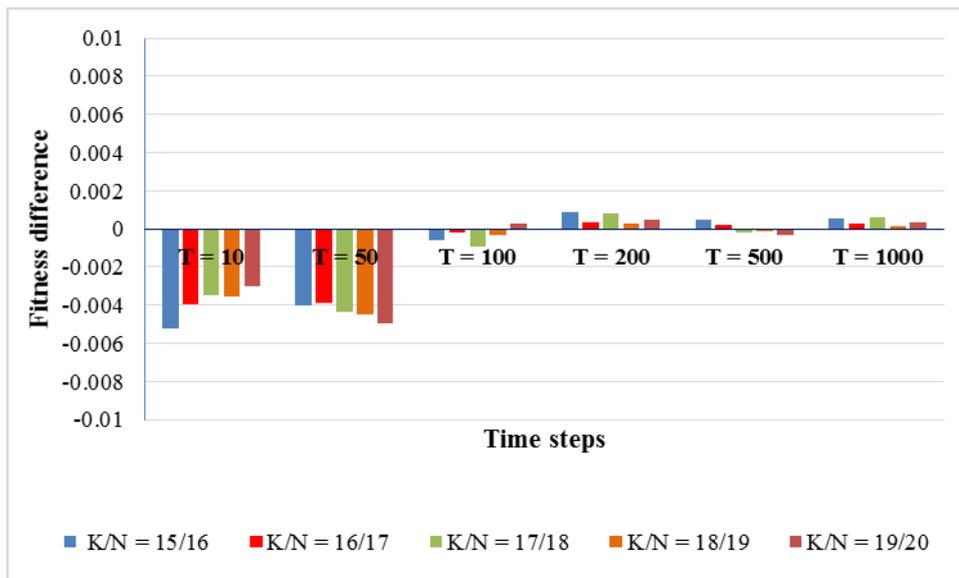

*Parameters.* Number of iterations = 10,000; *Fitness difference* = (overall fitness by **SMMLS** – overall fitness by **IMMLS**). Complexity is varied by varying **N** from 16 to 20 and using **K** = **N** -1.

In **Figure 8** I present the resource consumption picture for underlying the Figure 7 experiments for **IMMLS**. This set of graphs is analogous to the set of graphs presented in Figure 2. As before, the comparison is with **SMMLS**, which consumes 100% of the resources made available in the simulation experiments. The results shown in the two sets (Figure 2 and Figure 8) are similar in that the percentage of resources consumed as a percentage of the total resources available (i.e., the imposed limit of **T** computations of overall fitness of alternative decision-configurations) decreases as **T** increases. However the **T** = 50, **T** = 100, and **T** = 200 plots in Figure 8 show that when complexity in the sense **K/N** increases, the extent of resources consumed increases slightly. In contrast, in Figure 2 we noted what when interdependence (**K**) increases, resource consumption comes down. Notwithstanding the minor differences, when complexity is varied in highly uncorrelated landscapes (**K/N** is high), Figure 8 reconfirms that in comparison to the **SMMLS**, the **IMMLS** consumes significantly lower proportion of resources**.**

**Figure 8.** Resource consumption for **IMMLS** algorithm at high complexity

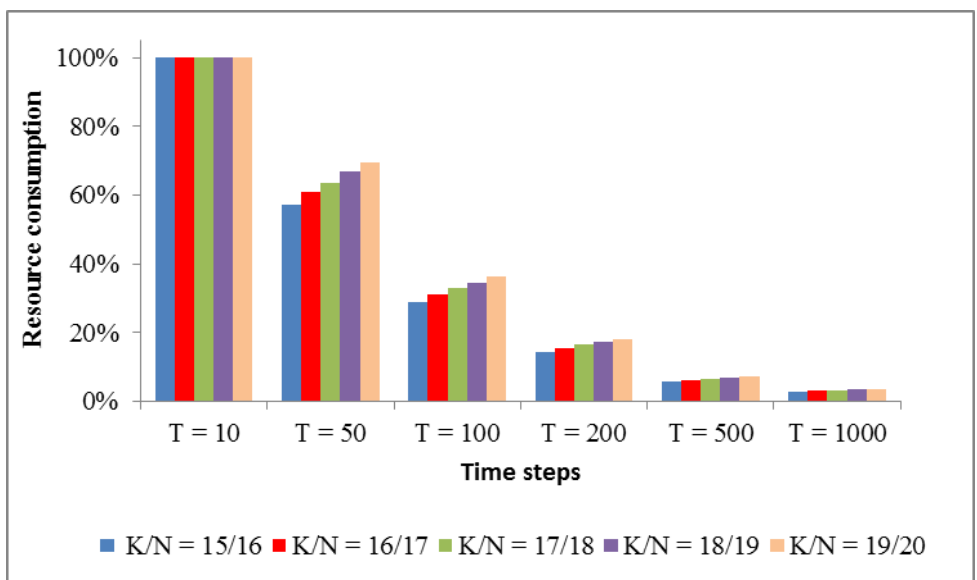

*Parameters.* Number of iterations = 10,000; ***Resource consumption*** represents the number of alternative configurations evaluated to arrive at the final decision-configuration. Complexity is varied by varying **N** from 16 to 20 and using **K** = **N** -1.

## 5   Discussion

I describe an algorithm (an *Intelligent Model of Myopic Local Search*, **IMMLS**) that allows search to logically terminate much earlier—saving computational resources—than the classical algorithm (the *Simple Model of Myopic Local Search,* **SMMLS**) in use for search for high-fitness decision-configurations on Kauffman's *NK*-landscape. Along the way, I provide a formal description of the process of **metering**—limiting the number of evaluations of alternative decision-configurations—to enable comparison across algorithms employing different logic to accomplish a common goal. I also propose a less resource intensive way of evaluating the **performance or payoff** outcome from an algorithm—by considering the ratio [(Final Overall Fitness – Initial Overall Fitness) / Initial Overall Fitness]—in a departure from prior approaches involving laborious calculation of overall fitness of all decision-configurations and reporting the ratio [Final Overall Fitness / Maximum Fitness attainable on a given *NK*-landscape].

   The **IMMLS** algorithm is superior to the classical **SMMLS** algorithm because it enables completing the search task with less number of computations, without endowing the algorithm with any additional ability over and above the abilities already possessed by the decision-maker agent in the classical algorithm. For example, in a given time step, the **IMMLS** algorithm carries out simple random sampling without replacement in evaluating single-node-flips in sequence. This requires remembering which nodes have already been flipped and which ones are still due. This ability already exists in the classical **SMMLS** algorithm: at the time of computation of overall fitness, the fitness contribution of individual nodes are taken up one at a time (following simple random sampling without replacement) in order to compute the average contribution across decision nodes for reporting overall fitness.

In comparison, certain researchers appear to have endowed **SMMLS** with additional capabilities—involving evaluating every neighboring configuration during fitness walk, but making a go/no-go decision only on the results of the first node flipped—in order to force the **SMMLS** to terminate early. For example, Bauman and Siggelkow [16] and Jain and Kogut [17] use **SMMLS** but make no mention of how they are making sure that the algorithm gets to know that a competency trap has been reached—an event upon which their respective algorithms change the way search is done. In such cases, switching to using the **IMMLS** algorithm can improve the quality of research.

**APPENDIX**: Figures equivalent to Figures 1 and 2 with **N** = 20 (instead of **N** = 16)

**Figure A1.** Difference in overall fitness between **SMMLS** and **IMMLS** for **N** = 20

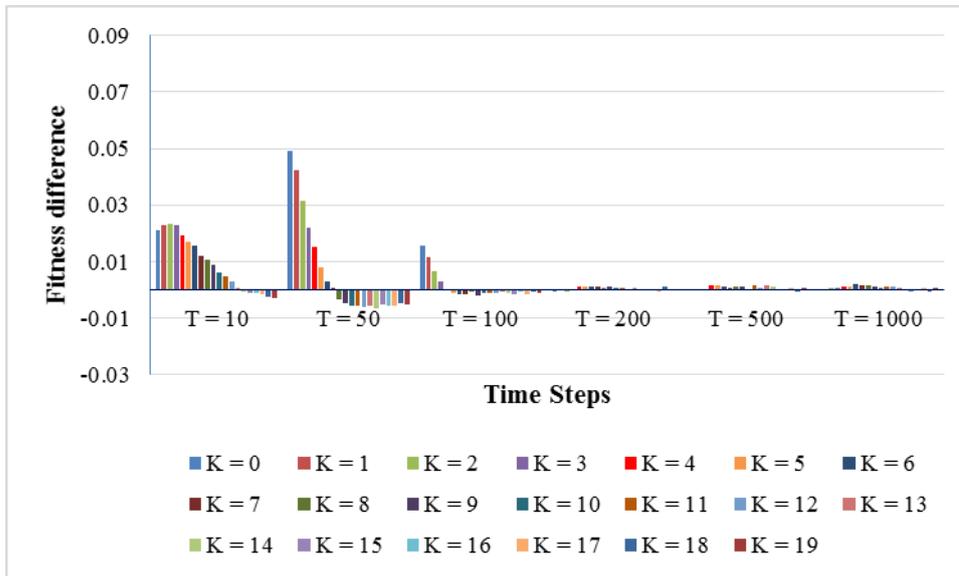

*Parameters.* **N** = 20; Number of iterations = 10,000; *Fitness difference* = (overall fitness by **SMMLS** – overall fitness by **IMMLS**).

**Figure A2.** Resource consumption for **IMMLS** algorithm for **N** = 20

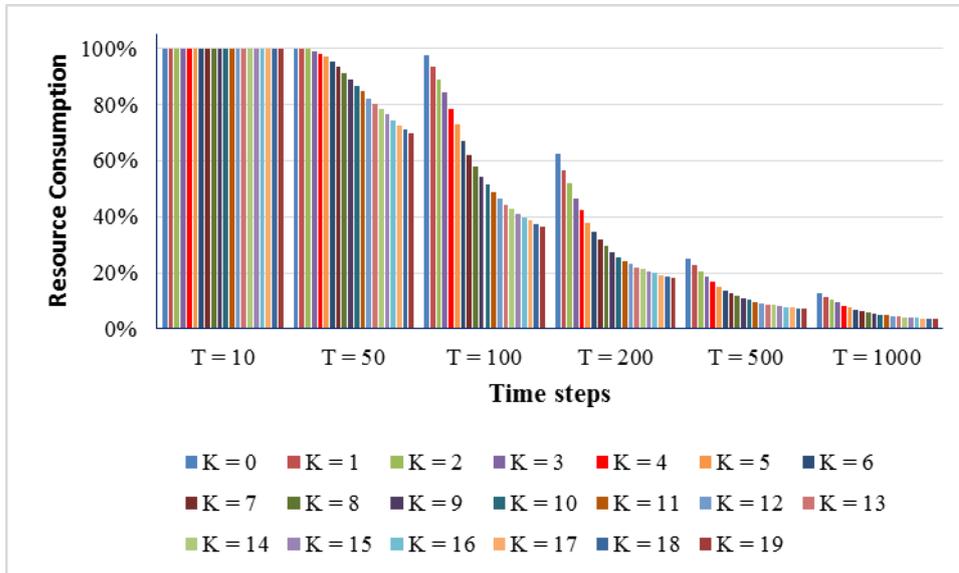

*Parameters.* **N** = 20; Number of iterations = 10,000; *Resource consumption* represents the number of alternative configurations evaluated to arrive at the final decision-configuration.